\documentclass[AMA,Times2COL]{WileyNJDv5}

\usepackage{graphicx}  
\usepackage{caption}
\usepackage{subcaption}
\usepackage{float}
\usepackage{array}
\usepackage{longtable}
\articletype{Research Article}%
\usepackage{float}

\startpage{1}
\begin{document}

\title{Non‐destructive Identification of Oyster Species is possible from Hyperspectral Images with Machine Learning}

\author[1,2]{Ethan K. Waters}

\author[1]{Max Wingfield}

\author[1]{Aiden Mellor}

\author[1]{Paul Stewart}

\author[1,3]{Iman Tahmasbian}

\authormark{Waters \textsc{et al.}}
\titlemark{Non‐destructive Identification of Oyster Species is possible from Hyperspectral Images with Machine Learning}

\address[1]{\orgdiv{Department of Primary Industries}, \orgaddress{\city{Toowoomba}, \state{QLD} 4350, \country{Australia}}}

\address[2]{\orgdiv{QUT Centre for Data Science, School of Mathematical Sciences, Queensland University of Technology}, \orgaddress{\city{Brisbane}, \state{QLD} 4000, \country{Australia}}}

\address[3]{\orgdiv{School of Environment and Science, Griffith University, Nathan}, \orgaddress{\city{Brisbane}, \state{QLD} 4111, \country{Australia}}}

\corres{Corresponding author Ethan Waters, \email{ethan.waters@dpi.qld.gov.au}}

\presentaddress{203 Tor St, Toowoomba City, QLD 4350}

\abstract[Abstract]{Differentiating between oyster species is important for developing new commercial oyster species suited to production systems and is critical in seafood supply chains for ensuring traceability or preventing food fraud. Common methods, such as DNA profiling, are destructive and time consuming. The possibility of using hyperspectral imaging (HSI) for discriminating between Black-Lip rock (BL) and Sydney rock (SR) oysters was investigated. Live BL and SR samples (N = 156) were scanned with a HSI camera (950-2515 nm). Partial Least Square Discriminant Analysis (PLS-DA) and Convolutional Neural Networks (CNN) were trained with Monte Carlo Cross Validation (MCCV) to distinguish BL and SR oysters from the spectral reflectance of their left and rights valves. The PLS-DA model successfully distinguished between the species from both the left and right valves with a median test set classification accuracy of 100\%, out performing the CNN with 83\% and 96\% respectively. Elemental and mineralogical composition in the surface and cross-section of oyster valves were measured with electron microscopy. Analysis of the right valve revealed that there was a greater number of layers in BL compared to SR (4 vs. 2). The concentrations of carbon and oxygen varied in the outer layer of the right valves, with BL being rich in carbon and SR being rich in oxygen. The variation in carbon and oxygen concentrations observed between BL and SR right valves may reflect differences in the relative abundance or composition of chitin and glycoproteins. This is supported by model-derived importance assigned to wavelengths associated with vibrational modes of functional groups characteristic of these compounds. However, transmittance analysis revealed that light was transmitted through the valves, especially around the valve edges, indicating that the spectral signatures may have been influenced by the other valve and/or the meat. Ultimately, the findings highlight an effective rapid, non-destructive methodology for oyster species identification with the potential to enable in-field species identification, improve operational efficiency in breeding programs and facilitates the potential use of wild spat as broodstock.}


\maketitle
\renewcommand\thefootnote{}
\footnotetext{\textbf{Abbreviations:} HSI, Hyperspectral Imaging; BL, Black-Lip Rock; SR, Sydney rock; PLS-DA, Partial Least Square Discriminant Analysis; DPI, Department of Primary Industries; CNN, Convolutional Neural Network}
\footnotetext{© The State of Queensland (through the Department Primary Industries) 2026.}
\renewcommand\thefootnote{\fnsymbol{footnote}}
\setcounter{footnote}{1}

\section{Introduction}\label{sec1}

Accurate differentiation among oyster species is fundamental to the sustainable development of aquaculture systems and the integrity of the seafood supply chains. Reliable species identification ensures hatcheries produce and distribute the intended species to maintain market traceability, product integrity, and biosecurity standards. Grower's husbandry and management decisions depend heavily on the correct species, as oyster species differ markedly in nutritional profiles, growth characteristics, environmental tolerances, potential value and disease susceptibility \citep{GuoOyster}. Cultivating inappropriate or disease-prone species can incur significant economic and biosecurity risks \citep{rubio2013onset}. Additionally, robust species discrimination supports government regulation, biosecurity surveillance, and selective breeding programs \citep{GuoOyster}. Morphological identification is often challenging or impractical \citep{GuoOyster, liu2021identification} and typically advanced analytical techniques, such as DNA profiling is relied upon \citep{GuoOyster}. However, DNA profiling is expensive, time consuming and requires a tissue sample of each individual organism which can lead to the death of potentially important broodstock. Consequently, DNA profiling is unsuitable for large-scale applications. Therefore, a non‐destructive technique that operates in real‐time is required to facilitate in-situ oyster identification in large-scale. 

Hyperspectral imaging (HSI) is a non-destructive, imaging-based form of Near-Infrared Spectroscopy (NIRS) that captures a continuous spectrum of reflected light, partitioning it into hundreds of narrow wavelength bands across the electromagnetic spectrum, allowing each sample to be characterised by a detailed spectral signature \citep{siesler2008near}. Depending on the physicochemical composition and structure of the object, light at different wavelengths is absorbed or reflected, which can provide insight into these factors. Consequently, these signatures can be analysed with machine learning algorithms to differentiate biological materials based on subtle chemical or structural differences. NIRS is now widely used across agriculture and aquaculture for a variety of tasks including nutrient monitoring \citep{brown2011, brown2012, guan2022}, phenotyping \citep{tahmasbian2025}, disease identification \citep{waters2025}, pest detection \citep{ahmad2018monitoring} and biomass estimation \citep{santos2015simple}. However, to the best of our knowledge, no published studies have investigated the use of HSI for oyster species identification. This work builds on recent developments in Queensland, Australia, where HSI successfully distinguished mud and saucer scallops for population survey applications \citep{tahmasbian2024}, showing the potential for non-destructive species discrimination in bivalves.

HSI has frequently been used in conjunction with machine learning and statistical modelling techniques to support informed decision-making. However, in many applied studies, the collection of large, well-balanced datasets is prohibitively expensive or logistically constrained. Consequently, Partial Least Square (PLS) methods remain widely adopted \citep{brown2011, brown2012, tahmasbian2025, chen2025machine} due to their robustness to multicollinearity, capacity to model high-dimensional data, and reliable performance in small-sample settings where the number of features is greater than or equal to the number of samples \citep{Aparicio21042011, wold2001pls}. A range of other machine learning approaches have been applied to HSI in aquaculture, with common choices including Random Forest, Extreme Gradient Boosting, Support Vector Machine and Convolutional Neural Networks (CNN) \citep{tahmasbian2025, chen2025machine, li2022predicting, kong2022rapid}. The choice and performance of these models varied depending on the specific problem constraints and dataset size.

This proof-of-concept study investigates whether different oyster species exhibit distinct spectral signatures that can be used for species identification at early growth stages with HSI. This study also investigated the mechanisms that might be repsonsible for HSI detecting differences between oysters using their spectral signatures.

\section{Methodology}\label{method}
\subsection{Sample Collection}\label{data_collection}
Two rock oyster species, Black-lip rock oyster (\textit{Saccostrea spathulata}, BL) [formerly Saccostrea echinata] and Sydney rock oyster (\textit{Saccostrea glomerata}, SR), were selected for HSI analysis. These species were chosen due to their strong visual similarity at early life stages and the importance of accurately differentiating between them for the Queensland oyster industry. Both species were bred and grown at the Bribie Island Research Centre (BIRC) in Queensland. The selected BL (N = 80) and SR (N = 76) oysters were 19 and 24 weeks of age, grown in the similar condition with no significant differences in environmental factors, including water chemistry, temperature and feed composition. The live oysters were transported in a cool box to the Department of Primary Industries (DPI) HSI laboratory in Toowoomba for analysis.

\subsection{Hyperspectral Image Acquisition}\label{HSI_Acquisition}
The HSI system used in this study was a line-scan shortwave infrared (SWIR) hyperspectral camera (HySpex SWIR-384, Norsk Elektro Optikk, Oslo, Norway) with a spatial resolution of 384 pixels per line. The camera measured spectral reflectance in the 950-2515 nm range, producing images with 288 spectral bands at a spectral resolution of 5.45 nm. Illumination was provided by two linear halogen light sources (100 W each), and samples were conveyed beneath the camera’s field of view with a motorised translation stage (Figure 1).

The samples were placed on the translation stage on one side and the exterior of both valves for each oyster were scanned across the spectral range (Figure 1). The process was repeated for the other side of the oysters. 

\begin{figure}[h]
    \centering
    \includegraphics[width=0.5\textwidth]{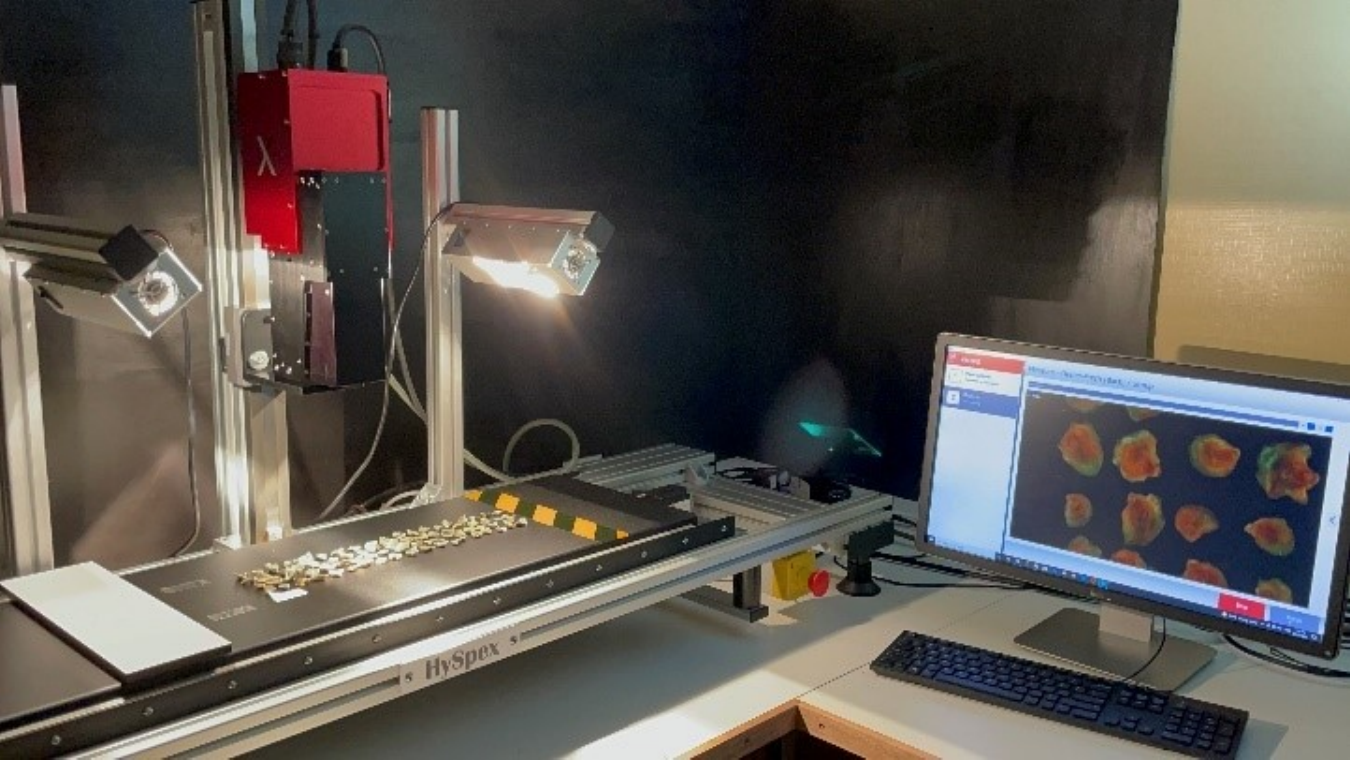}
    \caption{Hyperspectral imaging (HSI) system used to analyse oyster samples}
    \label{fig:camera}
\end{figure}

\subsection{Data Preprocessing}\label{data_preprocessing}
Preprocessing of hyperspectral data from the HySpex SWIR-384 first required radiometric calibration with black and white reference measurements to convert raw digital numbers into reflectance. White and dark reference measurements were used to account for illumination variability and sensor noise, respectively. The white reference was obtained by scanning a 50\% reflectance Zenith panel prior to sample acquisition, while the dark reference was captured by closing the lens. Reflectance was then computed according to Equation ~\ref{ref_eq} where R represented the corrected reflectance, $R_0$ was the raw image, $D$ was the dark reference, $W$ was the White image and $RT$ was the reflectance percentage of the standard target.

\begin{equation}\label{ref_eq}
R(\%) = \frac{R_0 - D}{W - D} \times R_T
\end{equation}

The background of the HSI data were identified and removed with a three-component PCA model, yielding a dataset of 312 individual oyster images of the left and right valves for all 156 oysters. For each oyster variety and valve perspective, pixel values within each spectral band were averaged, yielding a tabular dataset with 312 observations. 

\begin{figure}[h]
    \centering
    \includegraphics[width=0.5\textwidth]{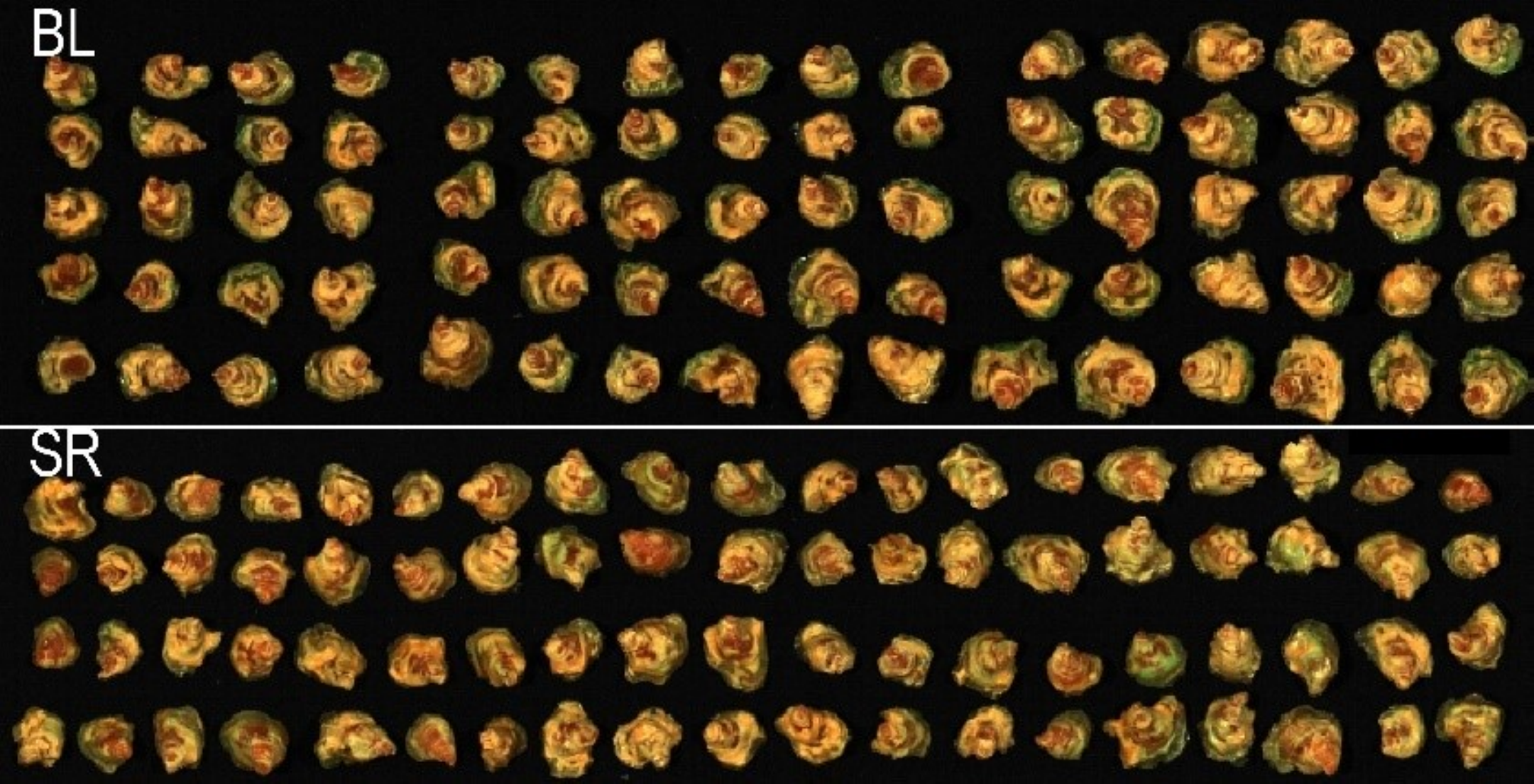}
    \caption{Pseudo-RGB image of the oysters used in this study, showing no distinct differences that can be used for discriminating the Black-lip rock (BL) and Sydney rock (SR) species. The average size (diameter) of BL and SR were 14.2 and 14.8 mm, respectively}
    \label{fig:rgb_oyster}
\end{figure}

Following hyperspectral preprocessing and background removal, pseudo RGB composite images were generated from the HSI data (Figure~\ref{fig:rgb_oyster}). These images illustrate that the live BL and SR oysters appear visually similar, with no readily distinguishable external features apparent to the human eye.

\begin{figure}[h]
    \centering
    \includegraphics[width=0.5\textwidth]{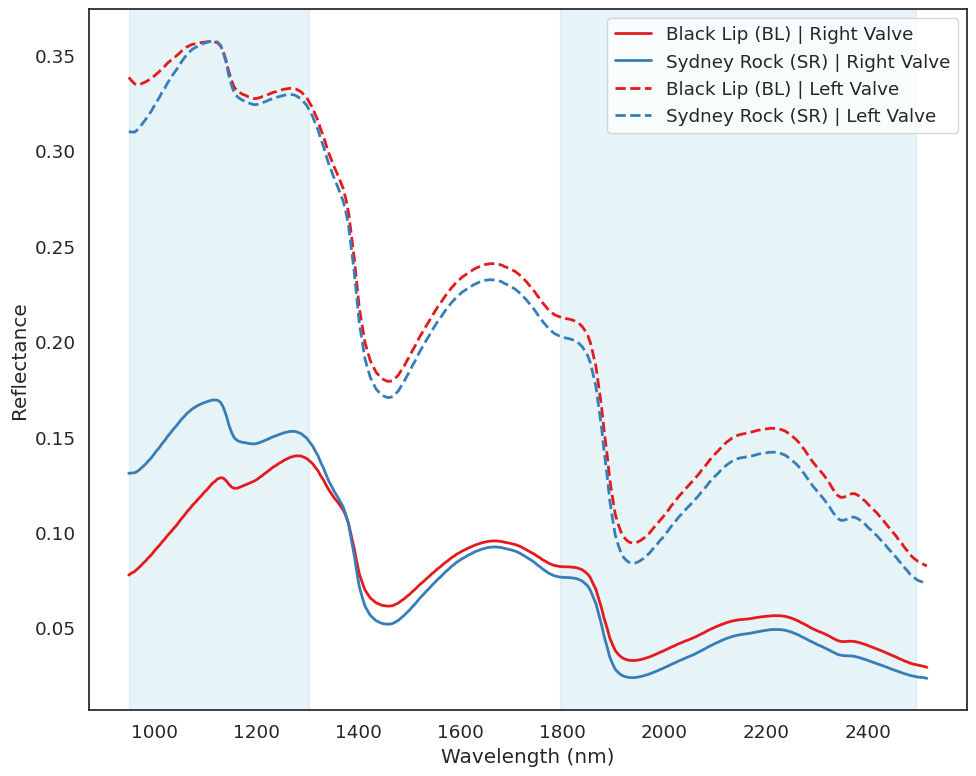}
    \caption{Average spectral reflectance of Black-lip rock (BL) and Sydney rock (SR) oysters, with regions showing a difference greater than 0.01 reflectance highlighted in light blue.}
    \label{fig:spectral_reflectance}
\end{figure}

An average spectral reflectance plot was generated from a training dataset for visualisation purposes and highlights the general spectral differences between species across the full wavelength range. The spectral signatures of live BL and SR oysters followed a broadly similar pattern. For a given valve, spectral responses were similar between species, while marked differences were evident between left and right valves (Figure~\ref{fig:spectral_reflectance}). Distinct differences were observed between approximately 950 nm and 1100 nm (left valve), and between 950 nm and 1300 nm (right valve). Across all wavelengths for both valves, the BL oyster generally exhibits higher reflectance than the SR oyster, with the exception of the shorter wavelengths between 950 nm and 1300 nm in the right valve. More subtle differences were observed between 1900nm and 2515nm for both valves. Typically across all wavelengths of the spectrum for both valves, the BL oyster reflects more light than the SR with  the exception of the short wavelengths 950nm and 1300nm right valve. 

\subsection{Species Classification}\label{data_classification}
HSI of the left and right oyster valves were analysed to compare classification performance. The procedure was carried out on the full spectral range and repeated with a reduced subset of wavelengths ($\leq$1050nm), which represents a more cost-effective configuration restricted to the range detectable by lower-cost silicon-based HSI cameras. This approach allows direct evaluation of whether comparable classification performance and feature importance can be achieved with a reduced-cost sensor configuration, thereby increasing the likelihood of adoption in industry.

\subsubsection{Model Selection \& Architecture}
Partial Least Squares Discriminant Analysis (PLS-DA) and CNNs was utilised to classify the oyster species. PLS-DA was selected as it is well suited for high-dimensional, collinear data where the number of samples (N=156) is less than the number of features (M=288) \citep{Aparicio21042011, wold2001pls}. PLS-DA seeks to identify a set of latent variables (LVs) that project the original predictors $\mathbf{X}$ into a lower-dimensional space, maximising covariance with the response matrix $\mathbf{Y}$, see Equation~\ref{plsda}. The weights, $\mathbf{W}$, define the projection of $\mathbf{X}$ onto the LVs, $\mathbf{P}$ are the $\mathbf{X}$-loadings, $\mathbf{Q}$ are the $\mathbf{Y}$-loadings, and $\mathbf{F}$ are the residuals ($\mathbf{F} = \mathbf{Y} - \hat{\mathbf{Y}}$). Increasing the number of LVs increases the model's capacity to capture complex patterns, but may lead to overfitting, while too few LVs can result in underfitting and suboptimal performance.

\begin{align}\label{plsda}
\mathbf{Y} = \mathbf{X W (P^\top W)^{-1} Q^\top + F}
\end{align}

CNNs have been widely recognised for their effectiveness in image-based classification tasks \citep{murphy2022probabilistic}. A CNN is organised as a sequence of convolutional, pooling, and fully connected layers that progressively transform raw pixel values into increasingly abstract feature representations. Convolutional layers slide learned kernels across the spatial dimensions of the input, computing a weighted sum at each position to produce feature maps, with each kernel capturing different local patterns such as edges, textures, or other distinctive structures. Pooling layers then downsample the spatial dimensions, reducing resolution while retaining the most salient activations. At the same time, convolutional layers can increase channel depth by applying multiple filters, allowing the network to represent more complex and abstract features. Consequently, as spatial resolution decreases, channel depth grows. Network parameters are optimised using mini-batch stochastic gradient descent (SGD), where at each iteration the weights $\theta$ are updated according to Equation~\ref{cnn_sgd}. Here, $\eta$ is the learning rate, $\mathcal{B}_t$ is the mini-batch at iteration $t$, and $\nabla_{\theta} \mathcal{L}(\theta_t; \mathcal{B}_t)$ denotes the gradient of the loss function with respect to the network parameters. Additionally, implementing a CNN allows the use of Gradient-weighted Class Activation Mapping (Grad-CAM++) to identify and visualise the spatial regions that contribute most strongly to oyster classification.

\begin{equation}\label{cnn_sgd}
\theta_{t+1} = \theta_t - \eta \, \nabla_{\theta} \mathcal{L}(\theta_t; \mathcal{B}_t)
\end{equation}

\begin{figure*}[htbp]
    \centering
    \includegraphics[width=0.8\textwidth]{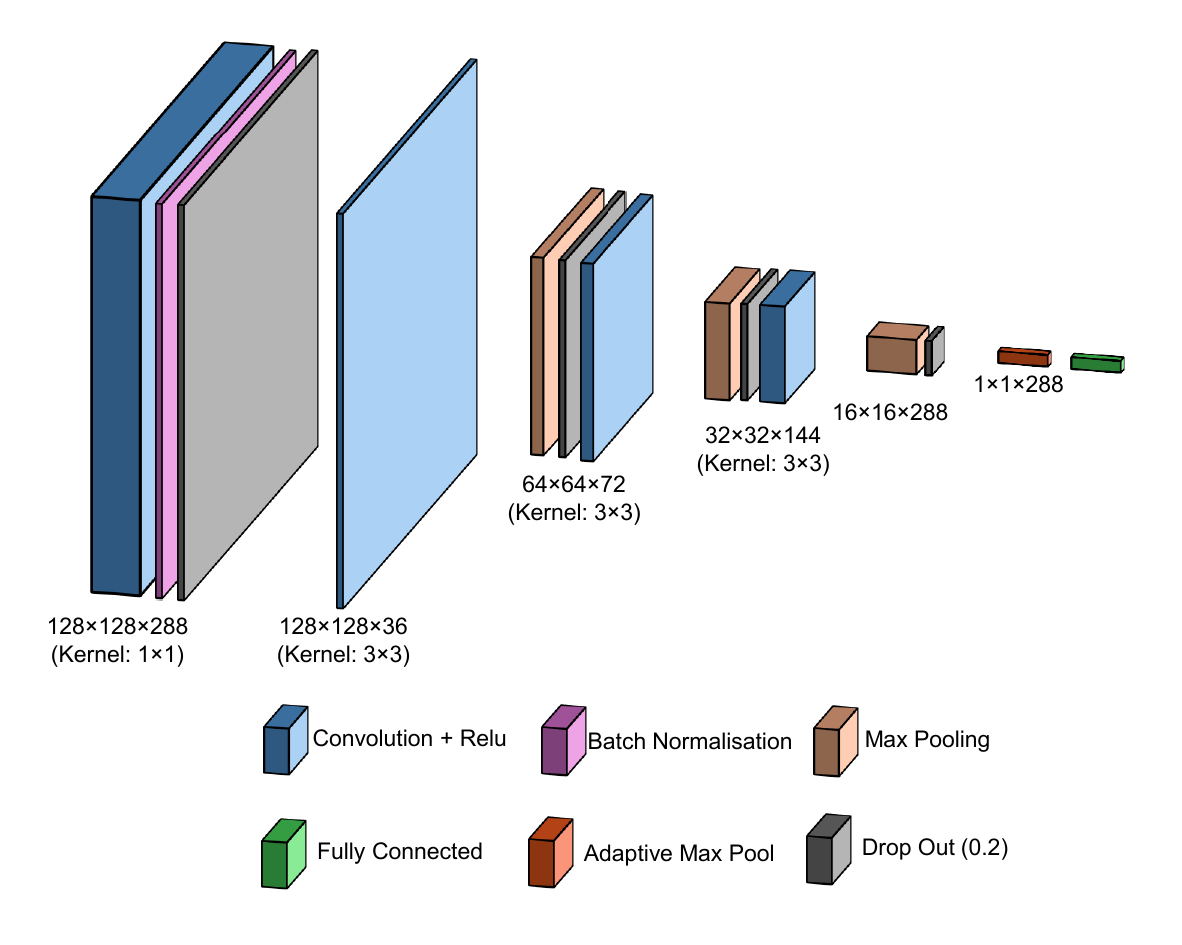}
    \caption{Architecture of the proposed OysterCNN model. The network begins with a 1×1 convolution applied to the input feature map to reduce the number of spectral features while preserving spatial resolution, followed by batch normalization and dropout. Subsequent layers use 3×3 convolutions to capture local spatial patterns. Rectified Linear Unit (ReLU) activations are applied after all convolutional layers. All max pooling was performed with a 2×2 kernel.}
    \label{fig:oystercnn}
\end{figure*}

The proposed OysterCNN architecture begins with a spectral reduction block that applies a 1×1 convolution to learn linear projections of the spectral dimension, compressing the input from 288 channels to 36, reducing computational cost while retaining the most informative spectral combinations. The subsequent spatial-spectral feature extraction block consists of three convolutional layers with progressive channel expansion and max-pooling, enabling the model to learn hierarchical spatial patterns. Batch normalisation and dropout layers were applied throughout the architecture for regularisation to mitigate the risk of overfitting. The proposed OysterCNN architecture can be seen in Figure~\ref{fig:oystercnn}.

\subsubsection{Model Training \& Validation}
Monte Carlo Cross Validation (MCCV) was selected as the preferred validation method over K-Fold Cross Validation due to its ability to provide more stable performance estimates with lower variance across repeated random splits of the data, albeit at the cost of increased computational complexity \citep{shan2022monte}. MCCV involves repeatedly generating random train-test splits of the dataset without replacement. In each iteration, the model is trained on the training set and evaluated on the test set, producing a distribution of performance metrics over the large number of iterations. In this study, 500 iterations were performed with a stratified 70-30 split to maintain class balance, providing a robust estimate of PLS-DA and CNN performance on unseen data. The following standard formulas were used to compute the metrics for each iteration, where $TP$ denotes true positives, $TN$ true negatives, $FP$ false positives, and $FN$ false negatives:

\begin{align}
\text{Accuracy} &= \frac{TP + TN}{TP + TN + FP + FN} \\
\text{Sensitivity (Recall)} &= \frac{TP}{TP + FN} \\
\text{Precision} &= \frac{TP}{TP + FP} 
\end{align}

The MCCV procedure was repeated across a range of LVs for the PLS-DA model, from 1 to 10 inclusive, to empirically identify the optimal number of LVs. The optimal number of LVs was determined as the smallest numbers of LVs that achieved the highest median classification accuracy across the 500 iterations. Hyperparameter tuning was not performed for the CNN model, as exhaustive search strategies in this way would be computationally prohibitive. Moreover, given the limited dataset size, extensive hyperparameter optimisation would increase the risk of overfitting to a small validation set \citep{cawley2010over}. To maximise the availability of training data while preserving model generalisability, literature-informed hyperparameter settings were adopted \citep{wojciuk2024improving, aszemi2019hyperparameter, bergstra2012random}. The full dataset was therefore retained for model training and evaluation.

\subsubsection{Feature Importance \& Selection}
The hyperspectral data contains a large number of highly collinear wavelengths, which can negatively impact model performance by introducing noise and redundancy. While PLS-DA inherently reduces dimensionality by projecting the data into a low-dimensional space, further refinement can be achieved by eliminating uninformative wavelengths. In this study, the Variable Importance in Projection (VIP) index was used to identify and remove wavelengths that contributed minimally to the model. VIP is a measure of relative importance of the predictors for all response variables in the PLS-DA model \citep{wold2001pls, wold1995pls}. It weights each predictor’s contribution to each LV by how much that LV explains the response variable, and then aggregates across LVs \citep{wold2001pls, wold1995pls}. Predictors with higher Variable Importance in Projection (VIP) values are considered more important, as they explain a greater proportion of the variance in the target variable. In this study, 0.8 was selected as the threshold to identify influential variables. The VIP index is calculated as follows:

\begin{align}
\text{VIP}_j = \sqrt{\frac{J \cdot \sum_{a=1}^{A} w_{ja}^2 , \text{SSY}_a}{A \cdot \text{SSY}_\text{total}}}
\end{align}

where $\text{VIP}_j$ is the importance of the jth wavelength, J is the total number of wavelengths, A is the number of LVs, $w_{ja}$ is the loading weight of the jth wavelength in the ath LV, $\text{SSY}_a$ is the explained sum of squares of the target variable in the ath LV, and $\text{SSY}_\text{total}$ is the total sum of squares of the target variable.

Within each MCCV iteration, the VIP index was calculated for the training set to determine wavelengths importance in that iteration. Wavelengths with VIP values below 0.8 were considered uninformative and removed from the model. The refined model, containing only the remaining informative wavelengths, was then evaluated on the corresponding unseen test set. VIP values were first averaged across iterations and then normalised to compare relative importance between models. 

Unlike PLS-DA, which requires manual feature extraction as a separate preprocessing step, the proposed OysterCNN architecture integrates spectral feature extraction and dimensionality reduction directly within the model through its initial 1x1 convolutional layer. To interpret model decisions within the integrated architecture, Grad-CAM++ was be used to generate class-discriminative spatial attribution maps for the CNN. Grad-CAM++ uses target class gradients to weight activations in the convoultional layers of a CNN to generate a heatmap that highlights key regions for predictions \citep{chattopadhay2018grad, selvaraju2017grad}. These complementary approaches offer different insights into the underlying factors driving model performance and feature importance.

\subsection{Software \& Implementation}\label{Software}

Scikit-learn’s PLSRegression module was used as the foundation and extended to implement PLS-DA for discriminant analysis. A custom class, PLSDAClassifier, was implemented, where the categorical class labels are one-hot encoded, and multivariate regression is performed on these one-hot encoded variables, with each class represented as a separate response variable. Class predictions were then obtained from the regression outputs by assigning each sample to the class with the highest predicted value.

This study was conducted with several software tools and libraries for data processing, analysis, and model development. Breeze version 2022.1.0 was used to capture, calibrate and process HSI data. Visual Studio Code version 1.100.2 was utilised as the programming IDE. Python 3.12.3 and a list of its libraries were utilised to conduct the analysis, including the following: matplotlib 3.10.3, seaborn 0.13.2, numpy 2.3.1, pandas 2.3.1, scikit-learn 1.7.0, scipy 1.16.0, spectral 0.24, and statsmodels 0.14.5.

\subsection{Laboratory analysis of oyster valves}\label{lab_analysis}
Inter-species variation in spectral reflectance was investigated through several laboratory analyses designed to evaluate whether morphological and compositional differences between oyster species contributed to the observed spectral patterns. These analyses focused on characterising physical structure, surface morphology, elemental composition, and organic content. A HSI system configured in transmittance mode was initially employed to determine whether light penetrated the valves, helping to establish whether the spectral signals primarily originated from surface features or internal layers.

To characterise external morphology, microscopy was used to assess height profile and surface roughness across 24 left valves, while visual inspection of both left and right valves from 12 individuals provided qualitative data on shape and colour variation. Structural mineralogy of both oysters was examined with X-ray diffraction (XRD) on the surface of both intact valves (inner and outer surfaces) and homogenised ground material of right valves from a subset of four oysters.

Elemental distribution and variation between species were assessed with X-ray fluorescence microscopy (XFM), performed on the inner and outer surfaces of 12 right valves and on the cross-section of a subset of four valves. To probe surface chemistry in greater detail, X-ray photoelectron spectroscopy (XPS) was conducted on the internal and external surfaces of four right valves, providing insights into the presence and relative abundance of surface elements. Complementary chemical characterisation was performed with Fourier Transform Infrared Spectroscopy with Attenuated Total Reflectance (FTIR-ATR) on ground samples from four right valves to detect molecular bonding patterns.

Structural features were further resolved through scanning electron microscopy (SEM), including backscattered electron (BSE) imaging and energy-dispersive X-ray spectroscopy (EDS), conducted on cross-sections of four right valves to visualise internal layering and map elemental composition. Finally, Desorption Electrospray Ionization (DESI) mass spectrometry was employed on the interior surfaces of the right valves to detect surface-bound organic compounds. Collectively, these techniques provided a multi-scale, multi-modal dataset to support interpretation of the spectral variation between oyster species.

\section{Results}
\subsection{Species Classification}
\begin{figure}[h]
    \centering
    \begin{subfigure}[t]{0.5\textwidth}
        \centering
        \includegraphics[width=\textwidth]{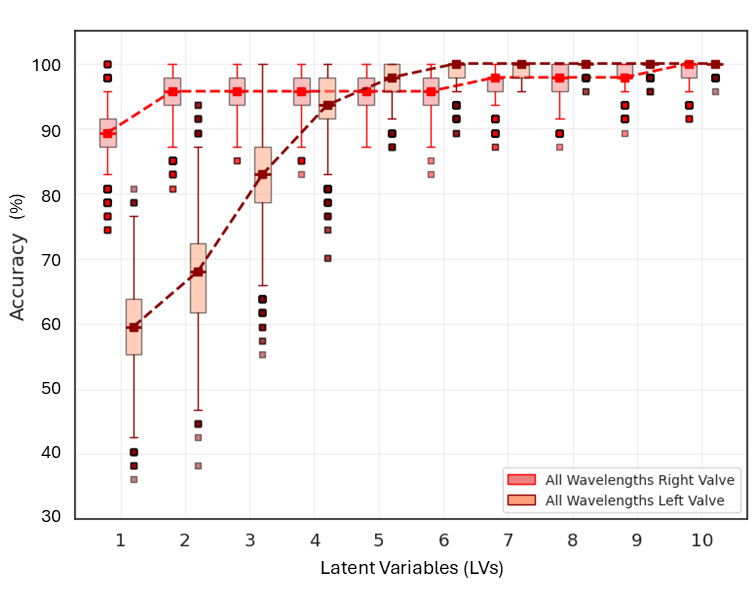}

        \label{fig:accuracy_full}
    \end{subfigure}
    
    \vspace{1em} 
    
    \begin{subfigure}[t]{0.5\textwidth}
        \centering
        \includegraphics[width=\textwidth]{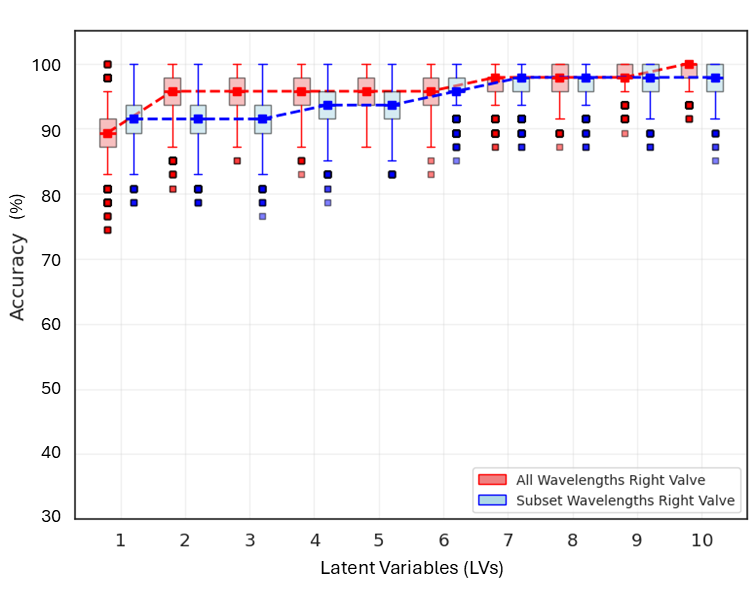}
        \label{fig:accuracy_subset}
    \end{subfigure}
    
    \caption{Distribution of accuracy for the Monte Carlo Cross Validation (MCCV) simulations of the PLS-DA model as the number of LVs increases. The top figure contrasts model performance for species classification from left and right valve images where all wavelengths were included as features (950-2515nm). While the bottom figure contrasts model performance for species classification from right valve images with a subset of wavelengths typically observed by more affordable HSI cameras (950-1100nm) and all available wavelengths.}
    \label{fig:accuracy_distributions}
\end{figure}

Both PLS-DA and CNN models were effectively able to classify oyster species from spectral reflectance data in the absence of visually distinguishable features. The models demonstrated strong performance on the validation sets with both the full feature set and the subset of wavelengths. The optimal number of LVs for the PLS-DA model varied depending on the available wavelengths and valve perspective (See Figure~\ref{fig:accuracy_distributions}). The model classifying species from HSI data of the left valve with all wavelengths achieved its highest median accuracy of 100\% with six LVs. Performance classifying oyster species based on the full wavelengths of the right valve required ten LVs to achieve its maximum median accuracy of 100\%. In contrast, classification from right valve images with a subset of wavelengths to imitate more cost-effective configurations reached its highest median accuracy of 97.9\% with seven LVs. Increasing the number of LVs beyond this resulted in the median accuracy plateauing, while variance increased, indicating potential overfitting with additional LVs.

\begin{figure}[h]
    \centering
    \begin{subfigure}[b]{0.45\textwidth}
        \centering
        \includegraphics[width=\textwidth]{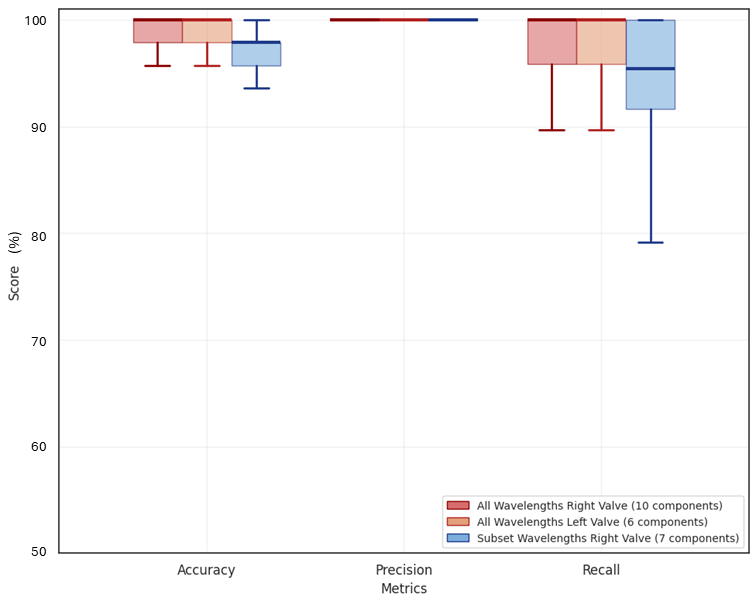}
        \caption{PLS-DA}
        \label{fig:final_performance_plsda}
    \end{subfigure}
    \hfill
    \begin{subfigure}[b]{0.45\textwidth}
        \centering
        \includegraphics[width=\textwidth]{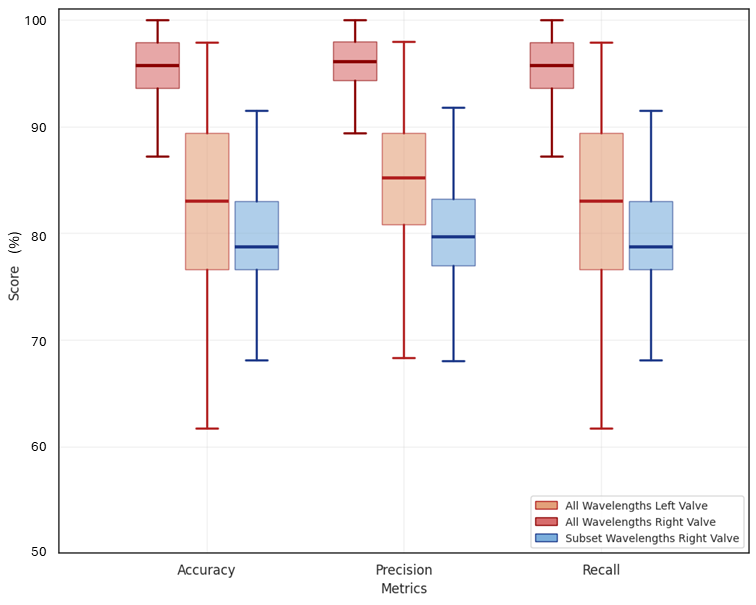}
        \caption{CNN}
        \label{fig:final_performance_cnn}
    \end{subfigure}
    \caption{Comparison of performance metrics for Partial Least Square Discriminant Analysis (PLS-DA) and Convolutional Neural Network (CNN) models across 500 Monte Carlo Cross Validation (MCCV) iterations. The PLS-DA subfigure illustrates the distribution of metrics for models with the optimal number of Latent Variables (LVs).}
    \label{fig:final_performance}
\end{figure}

For each dataset, the best-performing PLS-DA model with the least number of LVs was selected and further examined using additional metrics. All three models demonstrated comparable performance across all metrics for the different perspectives and available wavelengths (see Figure~\ref{fig:final_performance}). However, the model using a subset of wavelengths for the right valve performed slightly worse. While the models utilising all available wavelengths from images of the left and right valves showed very similar performance, the model for the left valve required fewer LVs to achieve this. Additionally, it was observed that recall exhibited greater variance compared to precision. The CNN models median accuracy was 83\%, 96\%, and 79\% for the left valve, right valve and wavelength subset from the right valve respectively. Performance across all metrics for the three HSI datasets and exhibited greater variance in performance than the PLS-DA models. Despite this, the CNN models for a given dataset produced similar performance distributions for recall, precision, and accuracy, indicating that the models were relatively well-balanced.

\begin{figure}[h!]
    \centering
    \includegraphics[width=0.5\textwidth]{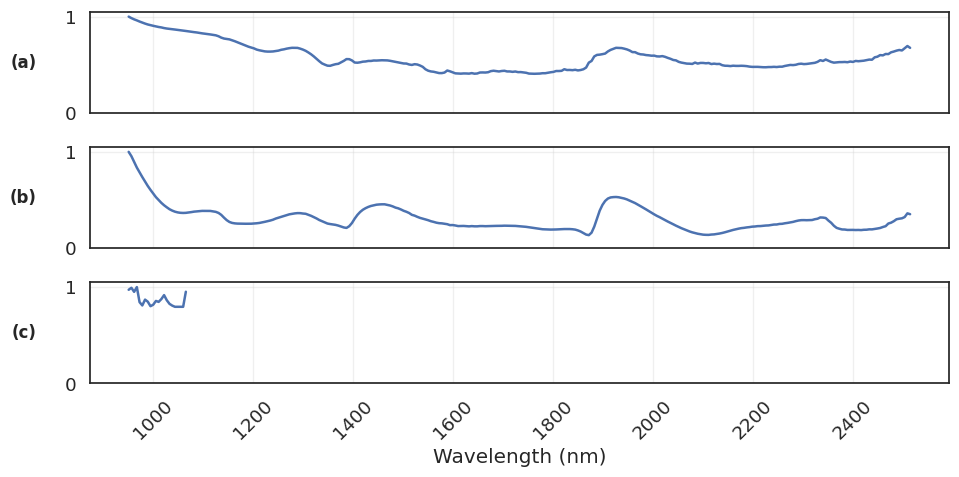}
    \caption{Normalised average Variable Importance in Projection (VIP) scores across 500 simulations, highlighting spectral regions most influential for classification. (a) Full-spectrum analysis of the right valve, (b) full-spectrum analysis of the left valve, and (c) reduced-spectrum analysis ($\leq$1050 nm) of the right valve, representing a lower-cost sensor configuration.}
    \label{fig:relative_vip}
\end{figure}

The importance of wavelengths varied depending on both the spectral range available and the valve perspective in the HSI (Figure~\ref{fig:relative_vip}). Across models, wavelengths below 1000nm were consistently identified as important, and, a notable region of importance occurred around 1900nm. Where all wavelengths were available, 98 wavelengths from the left valve and 143 wavelengths from the right valve were retained in all 500 iterations, while all 22 wavelengths in the reduced wavelength subset were retained in all 500 iterations. The ten most influential wavelengths across all model were generally in the range of 950-1000nm. The next 10 most important wavelengths (ranks 11-20) were generally located between 1900-1950nm for the left valve, but between 1000-1050nm for the right valve. While the 1900-1950 nm region was also identified as important for the right valve (Figure~\ref{fig:relative_vip}), a broader set of wavelengths contributed to the performance of models classifying species from the right valve, where more wavelengths considered important, in comparison to models classifying species from the left valve.

\begin{figure*}[!h]
    \centering

    \textbf{{Black Lip Left Valve}}
    
    \begin{subfigure}[b]{0.15\linewidth}
        \centering
        \includegraphics[width=\linewidth]{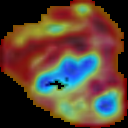}
    \end{subfigure}
    \begin{subfigure}[b]{0.15\textwidth}
        \centering
        \includegraphics[width=\linewidth]{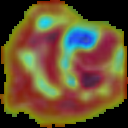}
    \end{subfigure}
    \begin{subfigure}[b]{0.15\textwidth}
        \centering
        \includegraphics[width=\linewidth]{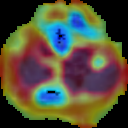}
    \end{subfigure}
    \begin{subfigure}[b]{0.15\textwidth}
        \centering
        \includegraphics[width=\linewidth]{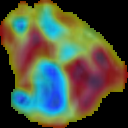}
    \end{subfigure}
    \begin{subfigure}[b]{0.15\textwidth}
        \centering
        \includegraphics[width=\linewidth]{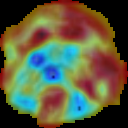}
    \end{subfigure}
    \begin{subfigure}[b]{0.15\textwidth}
        \centering
        \includegraphics[width=\linewidth]{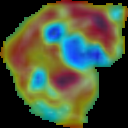}
    \end{subfigure}

    \textbf{{Black Lip Right Valve}}
    
    \begin{subfigure}[b]{0.15\textwidth}
        \centering
        \includegraphics[width=\linewidth]{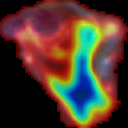}
    \end{subfigure}
    \begin{subfigure}[b]{0.15\textwidth}
        \centering
        \includegraphics[width=\linewidth]{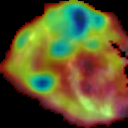}
    \end{subfigure}
    \begin{subfigure}[b]{0.15\textwidth}
        \centering
        \includegraphics[width=\linewidth]{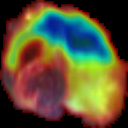}
    \end{subfigure}
    \begin{subfigure}[b]{0.15\textwidth}
        \centering
        \includegraphics[width=\linewidth]{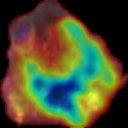}
    \end{subfigure}
    \begin{subfigure}[b]{0.15\textwidth}
        \centering
        \includegraphics[width=\linewidth]{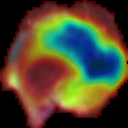}
    \end{subfigure}
    \begin{subfigure}[b]{0.15\textwidth}
        \centering
        \includegraphics[width=\linewidth]{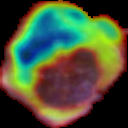}
    \end{subfigure}

    \textbf{{Sydney Rock Left Valve}}

    \begin{subfigure}[b]{0.15\linewidth}
        \centering
        \includegraphics[width=\linewidth]{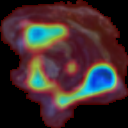}
    \end{subfigure}
    \begin{subfigure}[b]{0.15\textwidth}
        \centering
        \includegraphics[width=\linewidth]{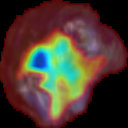}
    \end{subfigure}
    \begin{subfigure}[b]{0.15\textwidth}
        \centering
        \includegraphics[width=\linewidth]{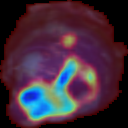}
    \end{subfigure}
    \begin{subfigure}[b]{0.15\textwidth}
        \centering
        \includegraphics[width=\linewidth]{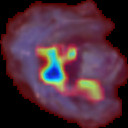}
    \end{subfigure}
    \begin{subfigure}[b]{0.15\textwidth}
        \centering
        \includegraphics[width=\linewidth]{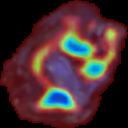}
    \end{subfigure}
    \begin{subfigure}[b]{0.15\textwidth}
        \centering
        \includegraphics[width=\linewidth]{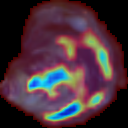}
    \end{subfigure}

    \textbf{{Sydney Rock Right Valve}}
    
    \begin{subfigure}[b]{0.15\textwidth}
        \centering
        \includegraphics[width=\linewidth]{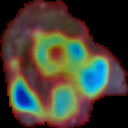}
    \end{subfigure}
    \begin{subfigure}[b]{0.15\textwidth}
        \centering
        \includegraphics[width=\linewidth]{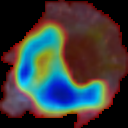}
    \end{subfigure}
    \begin{subfigure}[b]{0.15\textwidth}
        \centering
        \includegraphics[width=\linewidth]{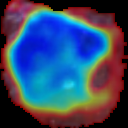}
    \end{subfigure}
    \begin{subfigure}[b]{0.15\textwidth}
        \centering
        \includegraphics[width=\linewidth]{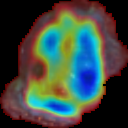}
    \end{subfigure}
    \begin{subfigure}[b]{0.15\textwidth}
        \centering
        \includegraphics[width=\linewidth]{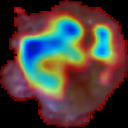}
    \end{subfigure}
    \begin{subfigure}[b]{0.15\textwidth}
        \centering
        \includegraphics[width=\linewidth]{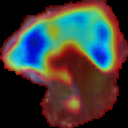}
    \end{subfigure}

    \caption{Gradient-weighted Class Activation Mapping (Grad-CAM++) visualizations for the left and right valves of Sydney Rock (SR) and Black Lip (BL) oysters. Heatmaps show the relative activations of gradients: red indicates regions of highest model attention, while blue indicates the lowest.}
    \label{fig:gradcam_oysters}
\end{figure*}

Grad-CAM++ indicated that the CNN attributed a larger spatial region of the SR oyster shell as important for classification compared with the BL oyster. Across samples, the regions identified as discriminative showed greater spatial variability in the left valve of the BL oyster than in the right valve. Areas most consistently highlighted by the model in right valve were located along the shell margins and around the hinge region (Figure~\ref{fig:gradcam_oysters}).

\subsection{Laboratory analysis of oyster valves}
Preliminary morphological assessment with optical microscopy revealed variation in surface topography among oyster specimens. Analysis of the left valves showed differences in height profiles both within and between species, however these were not statistically significant. BL oysters exhibited greater intra-species variability in surface roughness compared to SR oysters, whose profiles were more consistent across individuals. Visual inspection of valve morphology and coloration indicated no clear differences in the shape or appearance of left valves between species. However, the right valves of BL oysters appeared consistently darker than those of SR. These observations informed the subsequent structural and compositional analyses, which aimed to determine whether internal microstructure and elemental composition reflected these external morphological features.

Structural analysis with XRD revealed no major differences between BL and SR species in terms of surface calcite, aragonite, or cooperite content for both left and right valves. Similarly, ground right valve samples from both species exhibited comparable calcite diffraction patterns, and no cooperite was detected in ground form. However, the intensity of the aragonite diffraction signal was lower in BL right valve ground samples compared to SR, suggesting a potential difference in mineral crystallinity or abundance. Elemental analysis with XFM identified qualitative interspecies differences in the abundance of calcium, phosphorus, sulfur, zinc, chlorine, iron, copper, chromium, potassium, manganese, strontium, and titanium on valve surfaces. Similarly, the cross-sectional XFM analysis revealed slightly enriched phosphorus near the inner surface of BL right valves, distinguishing them from SR. Surface chemistry characterization with XPS confirmed the presence of calcium, carbon, oxygen, and silicon across all valve samples but could not confirm the presence of sulphur, iron or sodium. EDS identified calcium, oxygen, and carbon as the primary elements across all samples, with variable levels of magnesium, sodium, silicon, and sulfur. Notably, the outermost layers of BL right valves were enriched in carbon, while SR valves were oxygen-rich. 

\begin{figure}[h]
    \centering
   
    \begin{subfigure}[t]{0.23\textwidth}
        \centering
        \includegraphics[width=\textwidth]{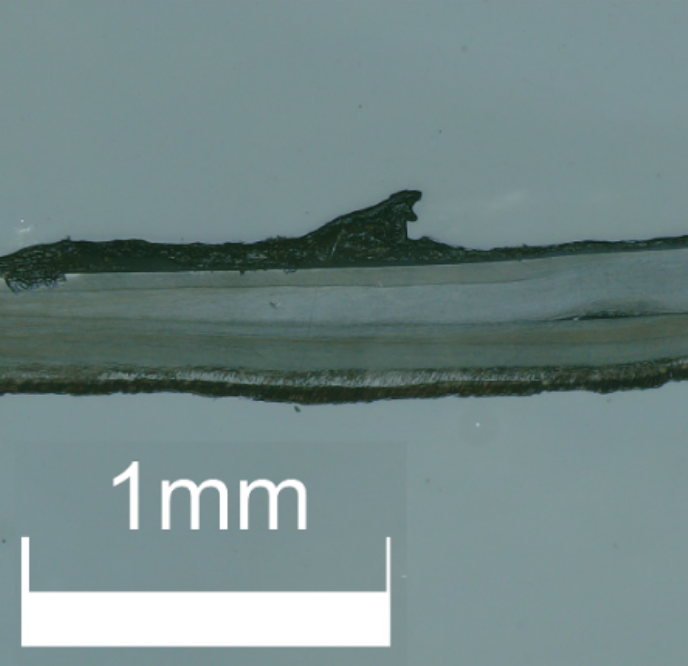} 
    \end{subfigure}
    \hfill
    \begin{subfigure}[t]{0.23\textwidth}
        \centering
        \includegraphics[width=\textwidth]{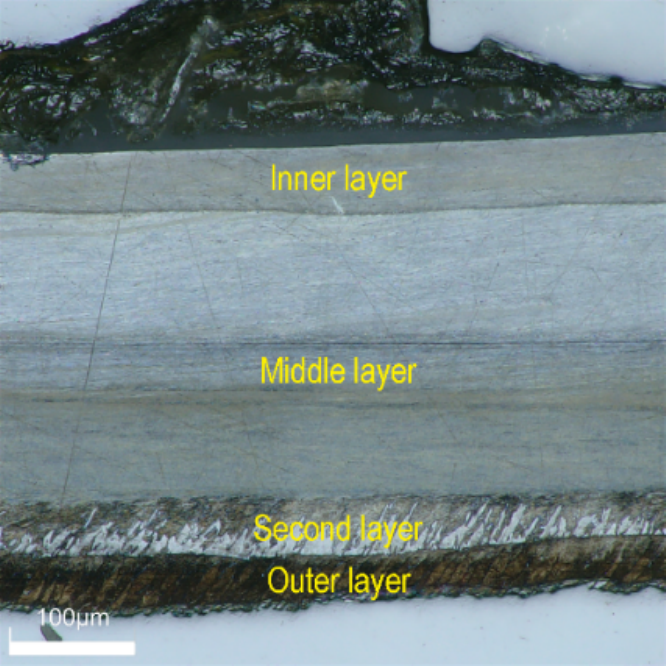} 
    \end{subfigure}
    
    \vspace{1em} 

    \begin{subfigure}[t]{0.23\textwidth}
        \centering
        \includegraphics[width=\textwidth]{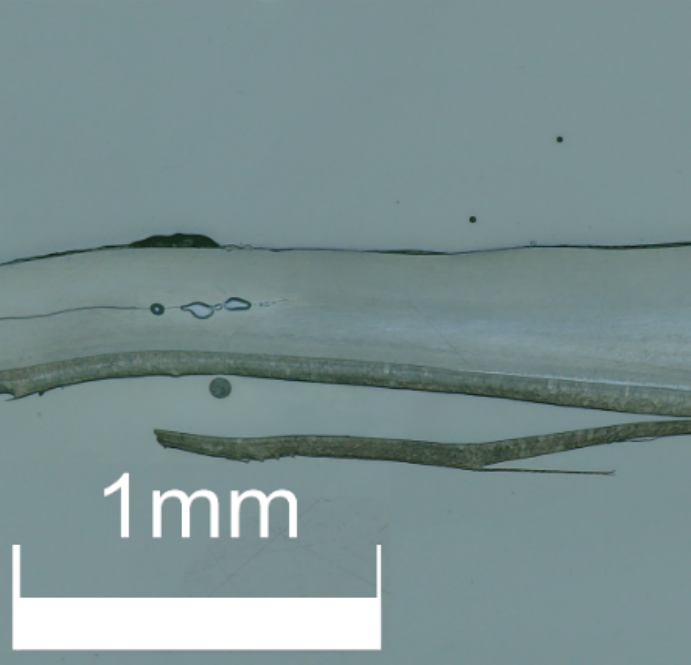} 
    \end{subfigure}
    \hfill
    \begin{subfigure}[t]{0.23\textwidth}
        \centering
        \includegraphics[width=\textwidth]{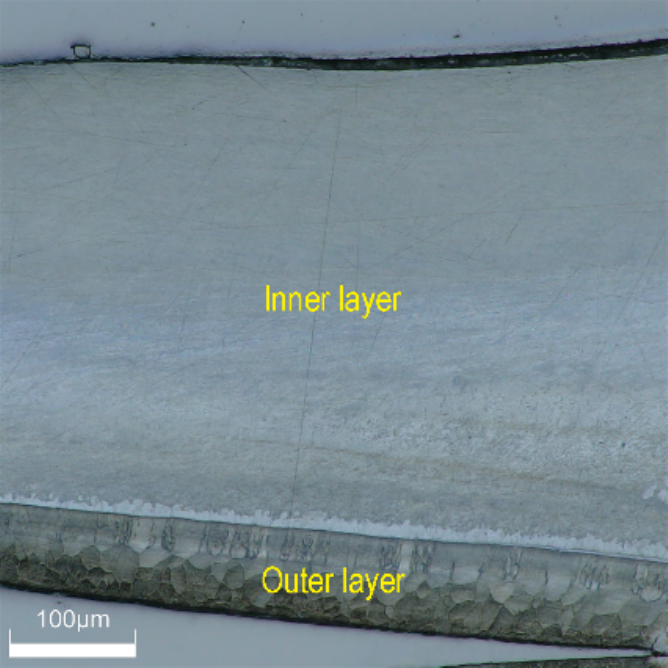} 
    \end{subfigure}

    \caption{Scanning electron microscopy of Black-Lip Rock (BL) (top) and Sydney Rock (SR) (bottom) oysters cross-section showing the different number of layers in the right valves}
    \label{fig:layers}
\end{figure}

SEM BSE revealed clear structural differences between species, where BL right valves consisted of four distinct layers, while SR valves contained only two (See Figure~\ref{fig:layers}). In both species, layer thickness varied along the valve circumference. Building on these structural observations, transmittance analysis confirmed that light transmission occurs through oyster valves in both species, with higher transmittance observed at the valve edges and clear spatial variation across the surface.

Finally, DESI indicated that no lipid were detected on the valves or any other significant variation in organic material between the BL and SR. A detailed summary of analytical techniques and corresponding results is provided in Appendix A.

\section{Discussion}
Despite their similar external appearance, this study demonstrated that PLS-DA and CNNs can accurately differentiate between BL and SR oysters from either the left or right valves from HSI. The achievement of 100\% median classification accuracy on the external test sets indicates that there are clear spectral differences between species which can be utilised for identification. Additionally, the median precision of species classification for the test sets was 100\%, indicating minimal misclassification of SR samples. Although the CNN models successfully classified oyster varieties, particularly when using the full wavelength set of the right valve images, their performance was not on par with the PLS-DA models. This discrepancy is likely attributable to the limited size of the dataset, as neural networks typically require large quantities of training data to achieve optimal performance. Nevertheless, the CNN models demonstrated a commendable level of accuracy given the constraints, highlighting their potential for further development with larger datasets. 

BL and SR oysters were largely comparable in valve composition, with the exception of differences in carbon and oxygen content, but exhibited clear differences in valve structure. Notably, the right valve of BL contained twice as many layers as SR (4 vs. 2). This increased layering provides a plausible structural basis for the divergent spectral reflectance patterns observed by HSI \citep{tolstoy2003handbook}. 

Oyster shells are primarily composed of calcium carbonate (CaCO$_3$), predominantly in the form of calcite and aragonite \citep{stenzel1963aragonite, ulagesan2022review}, with a minor organic matrix fraction (~0.1-5\%) consisting of polysaccharides such as chitin, and proteins, such as glycoproteins \citep{marie2012formation}. The observed variation in carbon and oxygen concentrations between BL and SR right valves may reflect differences in the relative abundance or composition of these organic constituents. Chitin and glycoproteins contribute functional groups such as N-H, O-H, C=O, and C-O, which influence spectral behaviour \citep{curran1989remote, siesler2008near}. Accordingly, wavelengths identified as important by VIP analysis aligned with vibrational modes associated with these functional groups, including O-H bending, first-overtone O-H stretching, and second-overtone N-H stretching (approximately 970-1050 nm), as well as O-H and C-O stretching and O-H deformation bands around 1900-1950 nm \citep{curran1989remote, siesler2008near}. These findings suggest that differences in organic matrix composition may underpin the observed spectral divergence between BL and SR oysters.

Relative gradient activations of the CNN highlight the edges and hinge areas of the right valve as potentially important for the identification of oyster species. These regions are typically associated with higher concentrations of the organic matrix \citep{marie2012formation}, supporting the idea that variations in the organic matrix composition, which influence carbon and oxygen concentrations, may underpin the observed spectral divergence between BL and SR oysters. However, transmittance testing showed that thin regions of the valve around the valve edges allowed light to pass through to varying degrees. These thin edges were characteristic of the young oysters used in this study (19-24 weeks old), whose shells had not yet fully thickened. Consequently, the spectral signatures may have been influenced by the underlying tissue and the other valve, which was not the primary focus. Collecting data from oysters at later growth stages, when shells are thicker and less transmissive, will be essential for determining whether species discrimination remains possible with a single valve. This is particularly important for applications in seafood supply chains where post-harvest species verification is required and in breeding programs where identification from the right valve alone is preferable due to its visibility. Furthermore, it is possible that the BL oysters exhibited fewer activations in the CNN due to the presence of additional layers, which may have caused greater light scattering. This increased scattering could reduce the importance of specific regions for species identification \citep{tolstoy2003handbook}.

In general, the left valve exhibited substantially higher spectral reflectance than the right valve across both oyster groups. However, due to cost constraints and the aim of maximising practical relevance for in-field application, laboratory analyses were conducted exclusively on the right valve. Consequently, direct comparisons of elemental or mineral composition between left and right valves, or between the left valves of BL and SR oysters, could not be undertaken and therefore cannot be inferred from the present results. Additionally, this limits our ability to speculate on the reasons behind the observed CNN activations for images of the left valve. Future studies should consider conducting laboratory analyses on both left and right valves to enable direct comparisons of elemental and mineral composition, which could provide deeper insights into the observed differences in spectral reflectance and their underlying causes.

An important practical finding was that a subset of wavelengths used to imitate a lower-cost spectrometer covering wavelengths between 950-1050 nm achieved comparable performance to the full HSI system for both left and right valves. This suggests that in-field, real-time identification may be feasible using portable, inexpensive devices. Although a subset of wavelengths was selected to mimic the capabilities of low-cost sensors, the high-grade HSI camera used in this study likely offers a superior signal-to-noise ratio within this wavelength range compared to cheaper sensors. As a result, the real-world performance of low-cost devices may not fully replicate the results observed here. However, these findings suggest that in-field assessment with a more affordable Visible and Near Infrared (VNIR) (400 - 1000 nm) sensor is plausible and could justify the associated expenditure. Future studies should aim to replicate this trial to confirm these findings under practical conditions.

In addition to replicating this trial under practical conditions, several other aspects could be addressed in subsequent studies. Wavelengths between 550-750 nm have been shown to be effective for non-invasive measurement of tissue oxygen saturation using spectroscopy \citep{franceschini1997, sanchez2023, stratonnikov2001, zuzak2002}. Given the oxygen-related differences observed in the outer layer between BL and SR, these wavelengths could provide particularly informative signals. Furthermore, this study focused exclusively on BL and SR, while many other oyster species are morphologically similar and typically require DNA profiling for accurate identification. Building on the strong discrimination achieved here, subsequent work could explore the potential of HSI to distinguish a broader range of oyster species. More broadly, future studies should expand the application of HSI to oyster health monitoring and early detection of diseases such as QX. Reliable non-destructive detection methods for QX would represent a significant milestone for the industry, given its devastating impact on oyster farming \citep{king2025sydney}. During QX outbreaks, it is difficult to distinguish between healthy and affected oysters among those that remain alive. Consequently, entire harvests may be deemed unmarketable despite the presence of potentially viable stock. It has been noted that live but QX-affected oysters may be distinguishable through the absence of a distinct growth edge on the shell \citep{king2025sydney}, suggesting that morphological indicators of shell growth could provide a potential avenue for discrimination. In this context, HSI offers a promising and underexplored approach for rapid nondestructive health assessment.

\section{Conclusion}
This study demonstrates that oyster species can be differentiated from SWIR hyperspectral imaging with partial least squares discriminant analysis and convolutional neural networks. The developed models achieved a high classification accuracy with a rapid, non-destructive methodology, highlighting a promising alternative to conventional destructive DNA-based identification. Importantly, this approach has the potential to enable in-field species identification, improve operational efficiency in breeding programs and facilitates the potential use of wild spat as broodstock. Additionally, it offers the potential for non-destructive species authentication within seafood supply chains at large scales, supporting traceability, labeling accuracy, and mitigation of food fraud. With further refinement through validation on larger and more diverse datasets, inclusion of additional oyster species, and assessment of lower-cost, field-deployable sensors, this approach has strong potential for scalable deployment and wider generalisability across aquaculture contexts.

\section*{CRediT authorship contribution statement}
\textbf{E Waters}: Conceptualization, Investigation, Methodology, Formal analysis, Validation, Software, Visualization, Data Curation, Writing - original draft, Writing - review \& editing. \textbf{A Mellor}: Conceptualization, Writing - original draft, Writing - review \& editing. \textbf{M Wingfield}: Conceptualization, Writing - original draft, Writing - review \& editing. \textbf{P Stewart}: Conceptualization, Resources, Writing - original draft, Writing - review \& editing. \textbf{I Tahmasbian}: Funding, Resources, acquisition, Conceptualization, Formal analysis, Methodology, Investigation, Data Curation,Visualization, Software, Project administration,  supervision, Writing - review \& editing

\section{Acknowledgments}
The authors sincerely thank the Technical Officers at the Department of Primary Industries Queensland, Bribie Island Research Station, for their dedicated efforts in caring for and collecting the samples. Additionally, the authors extended their gratitude to Andrew Norris, Kimberley Wockner and Rachele Osmond for their valuable feedback.

\section{Funding Information}
This work was funded internally by the Queensland Department of Primary Industries as an Innovation Project (AS12184).

\bibliography{references}
\clearpage
\onecolumn

\appendix
\begin{table*}[b]
\caption{Summary of laboratory analyses conducted on black-lip (BL) and Sydney rock (SR) valves}
\label{tab:lab_analyses}
\centering
\begin{tabular}{p{3cm}p{2.5cm}p{1cm}p{2.5cm}p{7.5cm}} 
\toprule
\textbf{Analysis} & \textbf{Method} & \textbf{N} & \textbf{Valve Sample Type} & \textbf{Details} \\
\midrule

Height profile and surface roughness & Optical microscopy & 24 & Left valve (solid surface) & Variations in height profile (left valve) were observed among individual samples between and within species. BL oysters exhibited more within-species variation, while SR oysters were more similar to each other. However, between-species variations in surface roughness and height profile were not statistically significant ($P > 0.05$). \\
\midrule
Morphology and colours & Visual variations & 12 & Left \& right (solid surface) & The shape and colour of the left valves were indistinguishable. However, the right valve of BL oysters was darker than that of SR oysters. \\
\midrule
Valve structure analysis & XRD & 4 & Left \& right (inside \& outside, solid surface) & Calcite, aragonite, and cooperite were observed with no distinguishable variation between BL and SR. The right valves were chosen for further studies due to colour differences. No differences were observed in the diffraction pattern of calcite. The intensity of the aragonite diffraction pattern was lower in BL compared to SR. No cooperite was detected in the ground samples. \\
\midrule
Elemental analysis and microscopy & XFM & 12 & Right (inside \& outside, solid surface) & Presence and qualitative variations in Ca, P, S, Zn, Cl, Fe, Cu, Cr, K, Mn, Sr, and Ti were observed within and between BL and SR samples. No distinguishable differences were observed between BL and SR in terms of nutrients. \\
\midrule
& XFM & 4 & Right (cross-section) & Slight differences in P distribution were observed between SR and BL. P appeared enriched at the inner surface for BL but at the outer surface for SR. However, the small sample size and limited sensitivity of XFM for P suggest these findings should be interpreted with caution. No differences were observed between BL and SR for other nutrients. \\
\midrule
& XPS & - & Right (inside \& outside, solid surface) & Ca, C, O, and Si were observed on the valve surface. The presence of S, Fe, Na, and other elements could not be confirmed. \\
\midrule
& FTIR-ATR & 4 & Right (ground) & No significant differences were observed in the FTIR-ATR spectra (4000 to 400 cm$^{-1}$) between BL and SR. \\
\midrule
& SEM BSE and SEM EDS & 4 & Right (cross-section) & All samples consisted of Ca, O, and C, with varying concentrations of Mg, Na, Si, and S. The number of layers differed between BL and SR valves (4 vs. 2). Layer thickness varied along the valve circumference for both types. The outer layer of BL valves was rich in C, while SR valves were rich in O, explaining the darker colour of the BL right valve. \\
\midrule
Organic compounds & DESI & - & Right (inside, solid surface) & No lipids were detected on the valves. No significant variations in organic material were observed between BL and SR. Due to large variations in valve height, DESI may misrepresent true variations and was deemed unreliable. \\
\bottomrule

\multicolumn{5}{p{16.5cm}}{\textit{Abbreviations:} X-Ray diffraction (XRD), X-ray fluorescence microscopy (XFM), X-ray photoelectron spectroscopy (XPS), Fourier transform infrared spectroscopy with attenuated total reflectance (FTIR-ATR), scanning electron microscopy-backscattered electron (SEM BSE), energy dispersive spectroscopy (EDS), desorption electrospray ionization (DESI), calcium (Ca), phosphorus (P), sulfur (S), zinc (Zn), chlorine (Cl), iron (Fe), copper (Cu), chromium (Cr), potassium (K), manganese (Mn), strontium (Sr), titanium (Ti), carbon (C), oxygen (O), silicon (Si), sodium (Na), and magnesium (Mg).} \\
\bottomrule
\end{tabular}
\end{table*}

\end{document}